%% file: MestavLuengoTong18ArxivFinal.tex
\documentclass[journal]{IEEEtran}
\usepackage{zref-base}
\usepackage{atveryend}
\makeatletter
\AfterLastShipout{%
	\if@filesw
	\begingroup
	\zref@setcurrent{default}{\the\value{equation}}%
	\zref@wrapper@immediate{%
		\zref@labelbyprops{eq-last}{default}%
	}%
	\endgroup
	\fi
}
\makeatother

\usepackage{cite}

\ifCLASSINFOpdf
\usepackage[pdftex]{graphicx}
\else
\usepackage[dvips]{graphicx}
\fi
\usepackage{amsmath}
\usepackage{algpseudocode}
\newsavebox{\ieeealgbox}

\usepackage{array}

\ifCLASSOPTIONcompsoc
\usepackage[caption=false,font=normalsize,labelfont=sf,textfont=sf]{subfig}
\else
\usepackage[caption=false,font=footnotesize]{subfig}

\usepackage{url}

\usepackage{algorithm}
\usepackage{mathrsfs}
\usepackage[mathscr]{eucal}
\usepackage{cite}
\usepackage{hyperref}

\usepackage{bbm}
\usepackage{chngpage}
\usepackage{color}
\usepackage{textcomp}

\usepackage{hyperref}
\usepackage{psfrag}
\usepackage{dsfont, pifont, times, import}
\usepackage{color}
\usepackage{makecell}
\usepackage{amssymb, amsmath, amsthm, enumitem}
\usepackage{mathtools}
\usepackage[nodisplayskipstretch]{setspace}
\usepackage[hyphenbreaks]{breakurl}
\usepackage[normalem]{ulem}
\usepackage{graphicx,amssymb,stfloats,multicol,cleveref,etoolbox}

\def\old#1{}    

\input LTmacros

\begin{document}
	
	\title{Bayesian State Estimation for Unobservable Distribution Systems via Deep Learning}
	
	\author{Kursat Rasim~Mestav,~\IEEEmembership{Member,~IEEE,}
		Jaime~Luengo-Rozas,~\IEEEmembership{Member,~IEEE,}
		and~Lang~Tong,~\IEEEmembership{Fellow,~IEEE}
		\thanks{Kursat Rasim Mestav, Jaime Luengo-Rozas, and Lang Tong are with the School of Electrical and Computer Engineering, Cornell University, Ithaca, NY, 14850 USA e-mail: {\tt \{krm264,jl3752,lt35\}@cornell.edu}.}%
		\thanks{This work was supported in part by the National Science Foundation under Awards 1809830 and 1816387, the Power System Engineering Research Center (PSERC), the Fulbright Scholar Program, and Iberdrola Foundation.}%
		\thanks{Part of the work was presented at 2018 IEEE Power  Energy Society General Meeting (PESGM) \cite{Mestav&Luengo-Rozas&Tong:18PESGM}.}
	}

	\maketitle
	
	\begin{abstract}
		The problem of state estimation for unobservable distribution systems is considered. A deep learning approach to Bayesian state estimation is proposed for real-time applications.  The proposed technique consists of distribution learning of stochastic power injection, a Monte Carlo technique for the training of a  deep neural network for state estimation, and a Bayesian bad-data detection and filtering algorithm. Structural characteristics of the deep neural networks are investigated.  Simulations illustrate the accuracy of Bayesian state estimation for unobservable systems and demonstrate the benefit of employing a deep neural network.  Numerical results show the robustness of Bayesian state estimation against modeling and estimation errors and the presence of bad and missing data.  Comparing with pseudo-measurement techniques, direct Bayesian state estimation via deep learning neural network outperforms existing benchmarks.
	\end{abstract}
	
	
	%
	\IEEEpeerreviewmaketitle
	
	\vspace{-0.1cm}

\section{Introduction}
\input introArxFinal

\section{Bayesian state estimation and neural network} \label{sec:II}
\input bayesArxFinal

\section{Learning with Deep Neural Network}\label{sec:III}

\input learningArxFinal

\section{Bad-Data Detection and Data Cleansing}\label{sec:IV}
\input badArxFinal

\section{Simulations Results and Discussions}\label{sec:V}
\input simulationArxFinal

\section{Conclusion}\label{sec:VI}

This paper presents a deep learning approach to Bayesian  state estimation for unobservable distribution systems under a stochastic generation and demand model.  The proposed approach employs two machine learning techniques: distribution learning of power injection and regression learning of MMSE estimator. A bad-data detection algorithm developed that detects and remove bad data prior to state estimation.   The use of deep neural network plays a  crucial role in overcoming computation complexity in Bayesian estimation, making the online computation significantly lower than the traditional WLS solutions. Simulation results demonstrate the potential of the Bayesian state estimation for cases that are otherwise intractable for conventional WLS-based techniques.

The Bayesian approach presented here has its limitations; further research and evaluations outside the scope of this paper are needed.  First, because prior distribution plays such an important role in Bayesian techniques, the estimator is less capable of adapting to changes in the network such as line and generation outages. Second, the training of deep neural networks is an area of research undergoing intensive investigation. Techniques available are largely ad hoc with little guarantee of performance.  For this reason, while the Bayesian approach is sound and promising, the performance reported here are results of careful tuning of some of the design parameters for the specific systems studied.

Several extensions are possible to improve further the proposed state estimation and bad-data detection techniques.    Although we focus on SCADA measurements, the approach developed here also applies to micro-PMU measurements that are not sufficient to achieve observability.  It is also possible to exploit temporal dependencies and employing convolutional or recurrent neural networks.
	
\section*{Appendix}\label{sec:VII}
\input appendixArxFinal

{
	\bibliographystyle{ieeetran}
	
	\bibliography{BIB}
}

\end{document}

%% file: LTmacros.tex
\def\nn{\nonumber}
\def\beq{\begin{equation}}
\def\eeq{\end{equation}}
\def\bea{\begin{eqnarray}}
\def\eea{\end{eqnarray}}
\def\ba{\begin{array}}
\def\ea{\end{array}}

\def\bitem{\begin{itemize}}
\def\eitem{\end{itemize}}
\def\ben{\begin{enumerate}}
\def\een{\end{enumerate}}

\def\eg{{\it e.g., \/}}

\def\ie{{\it i.e.,\ \/}}

\definecolor{bgrd}{rgb}{1,1,1}
\definecolor{gray}{rgb}{0.5,0.5,0.5}
\definecolor{dkr}{rgb}{0.7,0.1,0.2}
\definecolor{dkb}{rgb}{0.1,0.1,0.8}

\def\tcb{\textcolor{blue}}

\makeatletter
\newdimen{\captionwidth}
\long\def\@makecaption#1#2{%
\captionwidth .9\hsize
\vskip 10pt%
\setbox\@tempboxa\hbox{#1: #2}%
  \ifdim \wd\@tempboxa >\captionwidth%
    \setbox\@tempboxa\hbox{#1:\hspace*{.5em}}%
    \hfil\parbox{\captionwidth}{\raggedright\hangindent \wd\@tempboxa%
    \hangafter=1\unhbox\@tempboxa#2}\hfill%
  \else\centerline{\box\@tempboxa}%
  \fi
}
\makeatother

\def\edoc{\end{document}}

\def\span{\mbox{span}}

\def\T{\mbox{\tiny T}}

\newcommand{\mbbE}{\mathbb{E}}

\newcommand{\Smsc}{\mathscr{S}}

\newcommand{\Zmsc}{\mathscr{Z}}

\def\Hc{{\cal H}}

\def\Kc{{\cal K}}

\def\Nc{{\cal N}}

\def\Wc{{\cal W}}

%% file: introArxFinal.tex
We consider the problem of state estimation for power systems that have limited measurements.  We are motivated by the need for achieving a higher degree of situation awareness in distribution systems where the growing presence of distributed energy resources (DER) creates exciting opportunities and daunting challenges for system operators.  A compelling case can be made that effective state estimation is essential to optimize DER in real-time operations \cite{Lefebvre&Prevost&Lenoir:14PESGM}.

A major obstacle to  state estimation in distribution systems  is that such systems are nominally  {\em unobservable} \cite{Abur&Exposito:04book,Monticelli:12book}.   By unobservable it means that there is a manifold of uncountably many states that correspond to the same measurement.  System unobservability arises when the number of sensors is not sufficiently large---typical in distribution systems---or sensors are not well placed in the network.  An observable system may become unobservable when sensors are at fault, sensor data missing, or data tempered by malicious agents \cite{Kosut&Jia&Thomas&Tong:11TSG}.

A direct implication of unobservability is that the class of state estimators that
 assume  deterministic system state cannot provide  guarantees on the accuracy and consistency of their estimates.  In particular, the popular  weighted least-squares (WLS) estimator and its variants can no longer be used when the system is unobservable because small WLS error in model fitting does not imply small error in estimation; large estimation error may persist even in the absence of noise.

A standard remedy of unobservability is to use the so-called {\em pseudo measurements} based on interpolated observations or forecasts from historical data.  Indeed, the use of pseudo measurements has been a dominant theme for distribution system state estimation.    These techniques, however, are ad hoc and do not assure the quality of estimates.
More significantly,  historical data are often limited and have a poor temporal resolution for capturing real-time state dynamics.

The advent of smart meters and advanced metering infrastructure  provide new sources of measurements. Attempts have been made to incorporate  smart meter data  for state estimation \cite{Alimardani&etal:15TPS,WakeelWuJenkins:16AE,Gao&Yu:17IGST}.   Not intended for state estimation, smart meters measure  accumulative consumptions. They often arrive at a much slower timescale, \eg in 15-minute to hourly intervals, that is incompatible with the more rapid changes of DER. Unfortunately, existing techniques rarely address the mismatch of measurement resolution among the slow timescale smart meter data, the fast timescale real-time measurements (\eg  current magnitudes at feeders and substations), and the need of fast timescale state estimation.

State estimation for unobservable systems must incorporate additional properties beyond the measurement model defined by the power flow equations.  To this end, we pursue a  {\em Bayesian inference} approach where the system states (voltage phasors) and measurements are modeled as random variables endowed with (unknown) joint probability distributions.  Given the highly stochastic nature of the renewable injections, such a Bayesian model is both natural and appropriate.

The most important benefit of Bayesian inference is that  observability is no longer required.  A Bayesian estimator exploits probabilistic dependencies of the measurement variables on the system states;  it improves the prior distribution of the states using available measurements, even if there are only a few such measurements.  Unlike the least squares techniques that minimize {\em modeling error}, a Bayesian estimator minimizes directly the {\em estimation error}.

The advantage of Bayesian inference, however, comes with significant implementation issues.   First, the underlying joint distribution of the system states and measurements is unknown, and some type of learning is necessary.  Second, even if the relevant probability distribution is known or can be estimated, computing the actual state estimate is often intractable analytically and  prohibitive computationally.

\subsection{Summary of results and contributions}
As a significant departure from the pseudo-measurement approach to distribution system state estimation, this paper presents a novel application of Bayesian state estimation for possibly unobservable systems where  measurements may be unreliable, missing,  or subject to data attack.  Specifically, we develop a machine learning approach to the minimum mean-squared-error (MMSE) estimation of system states.  A pre-estimation bad-data detection and filtering algorithm based on the Bayesian model is also proposed.

The main benefit of the proposed Bayesian state estimation is twofold. First, the online computation cost of the Bayesian estimate is several orders of magnitude lower than that of the WLS techniques, thanks to the neural network implementation of the MMSE estimator.   Second,  the Bayesian inference framework provides a level of flexibility to assemble information from a variety of data sources.  For instance, smart meter data are not used directly in state estimation; they contribute to the learning of the probability distribution of network states.  The issues of incompatible timescales,  delayed measurements, and missing data are mitigated.

The proposed machine learning approach consists of {\em distribution learning} and {\em deep regression learning;}  the former uses smart meter data to learn bus injection distributions from which training samples are drawn. A novel contribution is learning the distribution of fast timescale power injection from slow timescale smart meter data. In regression learning, a deep neural network is trained for the MMSE state estimation.   A key innovation here  is the way that the power system model (the power flow equations) is embedded in the regression learning.


Numerical results demonstrate several features of the proposed approach. First, we show that the proposed Bayesian state estimator performs considerably better than the benchmark pseudo-measurement techniques, including those generating pseudo-measurements using neural networks. Second, we show that the proposed method is robust against inaccuracies in distribution learning\footnote{In our simulation, we generate test data based on a distribution learned in a different year.}. Third, our results suggest that using deep learning seems essential.  We observed that neural networks with five layers or more performed better than flatter networks, and neural networks with bulging middle sections performed better than rectangular ones. Finally, simulations show that the Bayesian bad-data detection and filtering is considerably more effective than the non-Bayesian pseudo-measurement techniques.

The proposed technique is of course not without limitations. Some of these limitations are summarized in Section \ref{sec:VI}.

\subsection{Related Work}
State estimation based on deterministic state models has been extensively studied.  See \cite{Abur&Exposito:04book,Monticelli:12book} and references therein.  We henceforth highlight only a subset of the literature with techniques suitable for distribution systems.

In some of the earliest contributions \cite{Roytelman&Shahidehpour:93TPS,Baran&Kelley:94TPS, Lu&Teng&Liu:94TPS,Ghosh&Lubkeman&Downey&Jones:97TPS}, it was well recognized that a critical challenge for distribution system state estimation is the  lack of observability.  Different from the Bayesian solution considered in this paper, most existing approaches are two-step solutions that produce  {\em pseudo measurements}  to make the system observable followed by applying WLS and other well-established techniques.

From an estimation theoretic perspective, generating pseudo measurements can be viewed as one of forecasting the real-time measurements based on historical data.  Thus the pseudo-measurement techniques are part of the so-called forecasting-aided state estimation \cite{Filho&DeSouza:09TPS,Filho&DeSouza&Freund:09TPS}.  To this end, machine learning techniques that have played significant roles in load forecasting can be tailored to produce pseudo measurements.   See, \eg \cite{Bernieri&Betta&Liguori&Losi:96TIM, Hippert&Pedreira&Souza:01TPS, Singh&Pal&Jabr:10IET,%
Manitsas&Singh&Pal&Strbac:12TPS, Wu&He&Jankins:13TPS, Onwuachumba&Wu&Musavi:13GTC,%
Adinolfi&Morini&Saviozzi&Silvestro:14PESISGT, Majeed&etal:14PESGM,%
Wang&etal:18TII, Zamzam&Fu&Sidiropoulos:18arXiv, Zhang&Wang&Giannakis:18arXiv}.

Bayesian approaches to state estimation are far less explored even though the idea was already proposed in the seminal work of Schweppe \cite{Schweppe&Wildes&Rom:70PAS}.
Bayesian state estimation generally requires the computation of the conditional statistics of the state variables.  An early contribution that modeled explicitly states as random was made in \cite{Ghosh&Lubkeman&Downey&Jones:97TPD} where load distributions were used to compute moments of states, although real-time measurements were used as  optimization constraints rather than as {\em conditioning variables} in Bayesian inference.    One approach to calculating conditional statistics is based on a graphical model of the distribution system from which belief propagation techniques are used to generate state estimates \cite{Hu&Kuh&Yang&Kavcic:11CIM,Chavali&Nehorai:15TSP}.  These techniques require a dependency graph of the system states and explicit forms of probability distributions.  Another approach is based on a linear approximation of the AC power flow \cite{Schenato&Barchi&Macii&Arghandeh&Poolla&Meier:14ICSGC}.

The approach presented in this paper belongs to the class of Monte Carlo techniques in which samples are generated and empirical conditional means computed. In our approach, instead of using Monte Carlo sampling to calculating the conditional mean directly as in \cite{Emami&Fernando&Iu&Trinh&Wong:15TPS, Angioni&Schlosser&Ponci&Monti:16TIM}, Monte Carlo sampling is used to train a neural network that, in real-time, computes the MMSE estimate directly from the measurements.

Bad-data detection and identification has been studied extensively \cite{Abur&Exposito:04book, Monticelli:12book}.
Classical methods are  {\em post-estimation} techniques where states are first estimated and used to compute the residue error.  The presence of bad data is declared if the  residue error exceeds a certain threshold. For such techniques, system observability is a prerequisite.  To identify and remove bad data, an iterative process is often used where state estimation is performed repeatedly after each positive bad-data detection and removal. Such techniques often fail to identify bad data or mistakenly remove good data.

In contrast to post-estimation bad-data detection, the method proposed in this paper belongs to the less explored class of {\em pre-estimation} detection and filtering techniques.  Several such techniques \cite{Falcao&Cook&Brameller:PAS82,Nishiya&Hasegawa&Koike:82IET,%
Abur&Keyhani&Bakhtiari:87TPS,Pignati&etal:14PSCC}
are based on exploiting a dynamic model to predicted the current measurement using past measurements, from which the prediction error becomes test statistics for bad-data detection.  In \cite{Salehfar&Zhao:95}, a neural network trained as an autoencoder\footnote{The authors of \cite{Salehfar&Zhao:95} did not use the autoencoder concept to explain their approach.}   is used to test against bad data.

%% file: bayesArxFinal.tex
\subsection{Algebraic and Statistical Models}
We adopt a generic static power flow model for an $N$-bus three-phase power system.  We assume that node 1 is the slack bus that, for distribution systems, represents the point of common coupling (PCC), where the distribution network is connected to the main grid.

The  three-phase voltage phasors at bus $i$ is a complex column vector  $x_i=[x_i^1,x_i^2, x_i^3]^\intercal $, where the superscripts are phase indices, and $x_i^k=V_i^k \angle\theta_i^k$, where $V_i^k$ is the voltage magnitude and $\theta_i^k$ is the phase angle for the state variable at phase $k$ of bus $i$. The overall system state $x=[x_1,\cdots, x_N]^\intercal$ is the column vector consisting of voltage phasors at all buses.

We adopt a static system  model defined by a pair of equations that characterize the relationship among the vector of (complex) power injections $s=[s_1,\cdots, s_N]^{\T}$ at network buses, the system states $x$, the vector of measurements $z=[z_1,\cdots, z_M]^{\T}$, and measurement noise $e$:
\begin{equation} \label{eq:pFlow}
x=g(s),~~z=h(x)+e,
\end{equation}
\noindent where $g(\cdot)$ is the mapping from net injection $s$ to the system state $x$ and
$h(x)$ is the measurement function defined by the sensor types and the locations in the network. Vector $z$ above includes standard types of measurements such as branch power flows, power injections, and current magnitudes.  Without loss of generality, we can treat the variables in (\ref{eq:pFlow}) as real by  taking either the rectangular or the polar form of the complex variables and modifying $g(\cdot)$ and $h(\cdot)$ in (\ref{eq:pFlow}) accordingly.   The specific forms and parameters of (\ref{eq:pFlow}) with different levels of modeling details can be found in \cite{Kersting:07book, Majumdar&Pal:16CMI}. In Appendix A functional relationship between some possible measurements and states is formulated.

Aside from the system model above, Bayesian estimation requires a probability model that specifies the statistical dependencies of the variables. Here we assume that the probability space is defined by the joint distribution of measurement noise $e$ and net power injection $s$.  We assume further that $e$ and $s$ are statistically independent.

\vspace{-0.5em}
\subsection{Bayesian State Estimation}
A state estimator $\hat{x}(\cdot)$ is a function of measurement $z$.  In Bayesian estimation, the estimator $\hat{x}(z)$ is defined by the joint distribution of   $x$ and  $z$ and the adopted optimization objective.  Here we focus on the minimum mean squared error (MMSE) estimator\footnote{The developed technique can be also applied to other Bayesian techniques such as the robust estimator based on the minimum absolute error (MMAE) estimator and the maximum aposteriori probability (MAP) estimator.} that minimizes the expected squared estimation error $\mbbE(||x(z)-x||^2)$ where the (Euclidean) 2-norm is used.

From the standard estimation theory, the MMSE estimator $\hat{x}^*(z)$ is given by the  mean of the state $x$ conditional on the realization of the  observation vector $z$:
\begin{equation}\label{eq:BSE}
\min_{\hat{x}(\cdot)} \mbbE(||x-\hat{x}(z)||^2)~~~\Rightarrow~~~\hat{x}^*(z)=\mbbE(x|z).
\end{equation}
Note the difference between the MMSE and the least squares estimators:
\[
\hat{x}_{\mbox{\tiny WLS}}(z)=\arg\min_x ||z-h(x)||^2,
\]
where the goal of least squares is to minimize the modeling error.

Simple as (\ref{eq:BSE}) may appear, the computation of the conditional mean can be exceedingly complex.  For all practical cases, the functional form of the conditional mean is unavailable.  More importantly, perhaps, the underlying joint distribution of $x$ and $z$ is unknown or impossible to specify, which makes the direct computation of $\hat{x}^*$ intractable.

%% file: learningArxFinal.tex
\subsection{Overview of Methodology}
\begin{figure}[h]
\begin{psfrags}
\psfrag{z}[c]{\small $z$}
\psfrag{xh}[c]{\small $\hat{x}(z)$}
\psfrag{w}[c]{\small $w$}
\psfrag{y}[c]{\small $y$}
\psfrag{F}[c]{\small $\hat{F}_s$}
\psfrag{s}[c]{\small $(s,e)$}
\psfrag{zx}[c]{\small $(x,z)$}
	\centering
	\includegraphics[width=1\columnwidth]{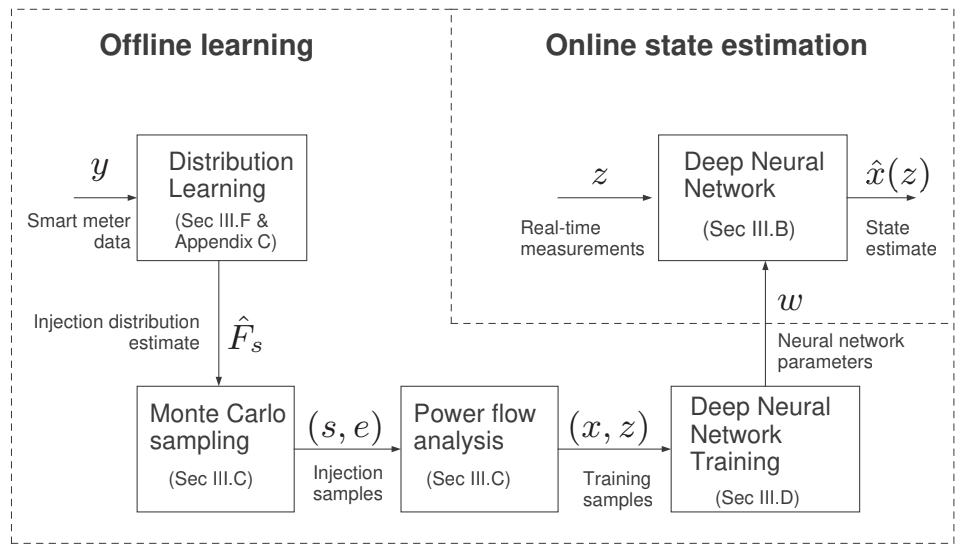}
	\hphantom{lol}
	\vspace{-0.8cm}
	\caption{\small Schematic of methodology.}
\label{fig:scheme}
\end{psfrags}
\end{figure}

We present an overview of the proposed methodology using Fig.~\ref{fig:scheme}.  Each functional block is explained in section(s) labeled.  All variables are consistent with those used in the paper.

The proposed scheme includes (i) online state estimation in the upper right of Fig.~\ref{fig:scheme} and (ii) offline learning in the rest of the figure.  This online-offline partition is of course not strict; the offline learning becomes online if learning continuous as data arrive.

The online state estimation is through a neural-network approximation of the MMSE estimator as described in Sec~\ref{sec:DNN}.   The offline learning includes distribution and regression learning modules.  Taking (historical) samples of  the smart meter measurement $y$, the distribution learning module produces $\hat{F}_s$ that approximates the  probability distribution of the net injection $s$.   The three submodules at the bottom of the figure are part of the regression learning that produces parameter $w$ of the neural network.  Specifically, the Monte Carlo sampling and the power flow analysis submodules generate samples from the estimated  net-injection probability distributions $\hat{F}_s$  and convert them to a set of  state-measurement training samples $\{(x,z)\}$.  The deep neural network training module sets the neural network parameter $w$ via an empirical risk minimization.

\subsection{Deep Neural Network Approximation} \label{sec:DNN}
The MMSE state estimator in (\ref{eq:BSE}) can be viewed as a {\em nonparametric regression} with the measurement $z$ as the regressor.  Such a regression is defined on the {\em unknown}  joint distribution of measurement $z$ and state $x$.   The  corresponding learning problem is of infinite dimensionality and intractable.

\begin{figure}[h]
	\centering
	\includegraphics[width=1\columnwidth]{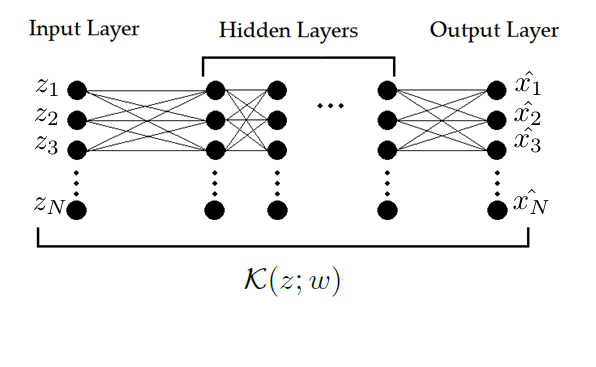}
	\hphantom{lol}
	\vspace{-0.8cm}
	\caption{\small Multi-layer Forward Neural Network.}
	\label{fig:NN2}
\end{figure}

We consider a finite dimensional approximation of (\ref{eq:BSE}) using a deep neural network shown in
 Fig~\ref{fig:NN2}.   The neural network consists of multiple layers of neurons. Neurons at each layer produce a vector output for the next layer using a (parameterized) nonlinear function of the output from the previous layer. The input-output relation $\Kc(\cdot; w)$ of the neural network is
\beq \label{eq:K0}
\hat{x}(z)=\Kc(z; w),
\eeq
where $w$ is the parameter matrix. For completeness, the specific form of $\Kc$ given in (\ref{eq:K0}) is developed in Appendix B.

The universal approximation theorem (see, \eg \cite{Sonoda&Murata:ACHA17}) has established that a neural network with a single hidden layer is sufficient to approximate an arbitrary continuous function. This means that with a sufficiently large neural network and appropriately chosen $w$, a neural network can well approximate the MMSE state estimator.   Under this approximation, the infinite dimensional learning problem of the conditional mean becomes a {\em finite dimensional} learning problem with dimensionality being the number of parameters in $w$.

\subsection{Regression Learning:  Training Samples}
We now focus on setting the neural network parameter $w$ to approximate  the MMSE state estimator $\hat{x}^*(z)$ in (\ref{eq:BSE}).
Standard deep learning algorithms apply to cases when there is a set of training samples made of the input-output pairs $\{(z, \hat{x}^*(z))\}$. Such pairs, unfortunately, are not available directly.  Nor do we have samples from which the underlying joint distribution $F_{x,z}$ of state $x$ and measurement $z$ can be learned directly.

The key to obtaining a training  set is to incorporate the underlying physical model characterized by model equation (\ref{eq:pFlow}). If we can learn the (marginal) distribution $F_s$ of the net power injection vector $s$, drawing an injection sample $s$ determines the state sample $x = g(s)$ from (\ref{eq:pFlow}). Assuming that independent measurement noise $e$ has distribution $F_e$,  $s \sim F_s$ and  $e \sim F_e$ produce a state sample $z=h(g(s))+e$.  Conceptually,

\beq \label{eq:se2xz}
(s,e) \sim F_s\times F_e \stackrel{{\small (\ref{eq:pFlow})}} {\longrightarrow} (x,z).
\eeq
Measurement noise distribution $F_e$ is assumed to be, say, zero-mean Gaussian $\Nc(0,\sigma^2)$.  The distribution of the net power injection, however, depends on a combination of load and possibly renewable generations.  We defer the discussion of  learning $F_s$ using data from the smart meter and other measurement devices to Section \ref{sec:IIIF}.

\subsection{Regression Learning:  Training Algorithm} \label{sec:IIID}
Given the set of training  samples
$$\Smsc=\{(x[k],z[k]), k=1,\cdots, |\Smsc|\}$$ generated according to (\ref{eq:se2xz}),   the weight matrix $w$ is chosen to minimize the {\em empirical risk} defined by
\begin{eqnarray}\label{NN_optimization}
L(w;\Smsc) &=& \frac{1}{|\Smsc|} \sum_{k: (x[k],z[k]) \in \Smsc} ||x[k] - \Kc(z[k];w)||^2,\nn\\
w^*&=&\arg\min_w  L(w;\Smsc). \nonumber
\end{eqnarray}
The empirical risk minimization problem above is well studied for deep learning problems, and an extensive literature exists.  See, \eg \cite{Goodfellow&Bengio&Courville:16book}.  For the state estimation problem at hand, the class of stochastic gradient descent algorithms is considered.  The Adam algorithm  \cite{Kingma&Ba:15ICLR} designed for non-stationary objectives and noisy measurements is particularly suitable.

A characteristic of deep learning is over-fitting, which means that the number of neural network parameters tends to be large relative to the training data set. A general approach to overcoming over-fitting is regularization that constraints in some way the search process in neural network training. Standard techniques include $L_{1}$ regularization, dropout, and early stopping  \cite{Goodfellow&Bengio&Courville:16book}. We present next a regularization technique based on structural constraints on the neural network.

\subsection{Regression Learning:  Neural Network Structure} \label{sec:IIIE}
The performance of state estimation can be affected by the structure of the neural network.    The ``shape'' of the network  can vary from shallow-and-wide  to  deep-and-narrow, and it does not have to be rectangular. Indeed, the shape of the network plays a role or regularization.

We propose a technique that transforms a rectangular network to a non-rectangular one based on the statistical clustering of the output variables at each layer.   The intuition is that, for a feedforward network, if the outputs of two neurons at the same layer are strongly correlated statistically, they can be combined.   Generalizing beyond two neurons, if the outputs of a group of neurons are highly correlated, this group of neurons may be well represented by a single neuron.

\begin{figure}[h]
	\centering
	\includegraphics[width=1\columnwidth]{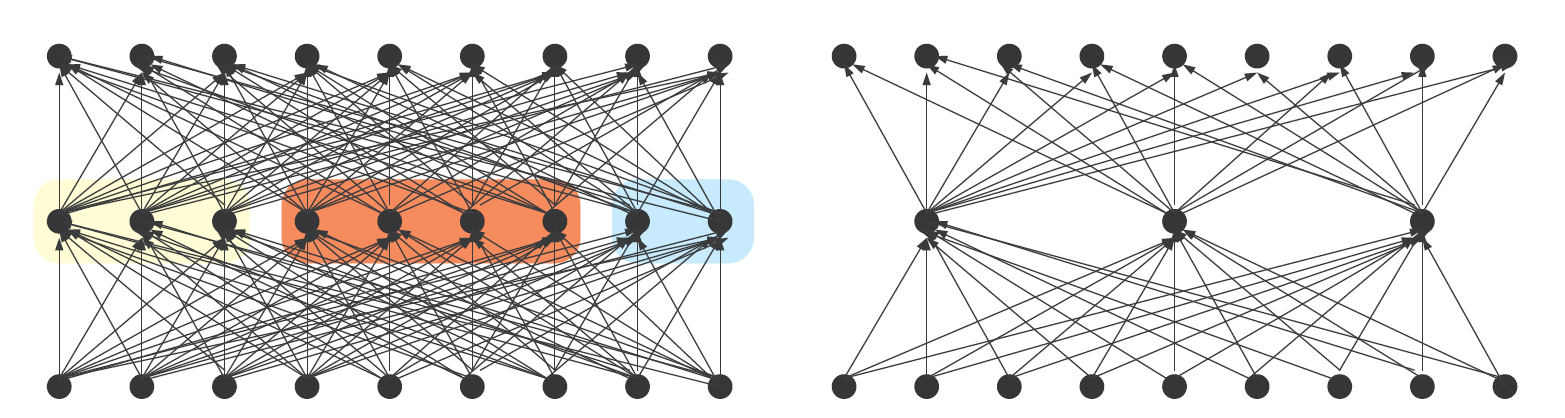}
	\hphantom{lol}
	\vspace{-0.8cm}
	\caption{\small A reduction of neurons in a layer through clustering.  Left panel: the right three, the middle four, and right three form three clusters.  Right panel: each cluster is represented by a single neuron to form a reduced network.}
	\label{fig:clustering}
\end{figure}

As shown in Fig.~\ref{fig:clustering}, if neurons at a particular layer  are clustered into three groups for the network in the left panel, the network on the left panel is reduced to one on the right, replacing each subgroup of neurons by its single-neuron representative.

Clustering requires a similarity measure. To this end, we considered a standard similarity measure $\rho_{X,Y}$ between random variables $X$ and $Y$ defined by
\[
\rho_{X,Y} = 1-\frac{\mbbE(XY)}{\sqrt{\mbbE(X^2)\mbbE(Y^2)}}.
\]
A variety of clustering techniques can be applied once a similarity measure is chosen.  In particular,  hierarchical agglomerative clustering \cite{Day:Class84} allows us to control the number and the sizes of clusters.

In Section \ref{sec:V}, we present numerical results on the structure of the neural network using the above clustering analysis.  The results from these experiments suggest that a network with a bulging middle section seems to perform better.

\vspace{-0.5em}
\subsection{Learning Net Injection Distributions}  \label{sec:IIIF}
The distribution of net injection needs to be learned to generate training samples.
Distribution learning can be parametric or non-parametric \cite{Wasserman:book1, Wasserman:book2}. By restricting the distribution class, parametric techniques are well developed. The assumption on the parametric class can be wrong, however, in which case very little can be said about the performance. This makes it highly desirable that the Bayesian estimator is robust against errors in distribution learning.

Because of the historical data for injections are limited, we estimate the injection distribution based on the parametric model of Gaussian mixtures commonly used to model load and renewable generations  \cite{Manitsas&Singh&Pal&Strbac:08SGD, Singh&Pal&Jabr:10TPS, Sanjari&Gooi:17TPS 
}.  The maximum likelihood (EM) method is used in the estimation.

If power injection can be measured locally at injection points, the injection distribution can be estimated locally without communicating the data to the central location; only the distribution parameters need to be transmitted. However, if measurements of net injection are not available, we propose an alternative technique that uses smart meter data to estimate the power injection distributions. This is nontrivial because smart meters typically measure accumulative consumptions, and smart meter data are collected at a much slower timescale. In Appendix C, we present a time-series based technique that exploits the underlying structure of the Gaussian mixture distributions. This technique converts the Gaussian mixture parameters of the smart meter data distributions to distribution parameters for the fast timescale power injections.

\subsection{Computation complexity}
The computation costs of the proposed Bayesian estimator include the cost of online calculation of state estimates and that of offline learning.
 Specifically, to estimate an $N$-dimensional state vector with a neural network of fixed depth requires roughly $O(N^2)$ computations.   Special hardware  that exploits massive parallelism can greatly speed up the computation \cite{Sze&Chen&Yang&Emer:17Proc}.   In contrast, standard second-order techniques for the WLS estimator  have the  cost  of $O(N^3)$ per iteration.

The computation cost of offline training is more difficult to quantify; it depends on the size of training data set,  the algorithm used for training, and the number of iterations required to achieve some level of accuracy.    Note that, for the WLS method, there is a negligible cost in obtaining a measurement sample whereas, for the Bayesian state estimation, the cost of Monte Carlo sampling is nontrivial.

For the proposed algorithm, generating a single sample requires solving the power flow equation. A standard implementation of Newton Ralphson technique requires the evaluation of the inverse of the Jacobian matrix with the (per sample) cost  roughly of the order $O(N^3)$.  It seems necessary that at least $O(N)$ training samples are generated, the cost of generating samples is roughly $O(N^4)$.  For the stochastic gradient descent algorithm used for training, the computation cost of a neural network with fixed depth is the cost of evaluating the gradient, which is roughly $O(N^2)$.

In summary, ignoring the cost associated with iteration,   we see that the online computation cost of the proposed Bayesian state estimation is considerably lower than the WLS methods, $O(N^2)$ vs. $O(N^3)$.  On the other hand, if the Bayesian approach is implemented online, the cost is considerably greater than that of the  WLS algorithms,    $O(N^4)$ vs. $O(N^3)$.

%% file: badArxFinal.tex
State estimation relies on data, and the quality of data depends on the types of sensors, the quality of the collection and  communications, and the possibility of cyber data attacks where measurement data are manipulated by an attacker\cite{Kosut&Jia&Thomas&Tong:11TSG}, \cite{Kim&Tong&Thomas:15TSP}. Bad-data detection and mitigation are an integral part of power system state estimation.

We first consider the simpler case of missing data that may be  results of  packet drops, communication delays, or that the data are deemed corrupted  and removed.  While no detection is needed in this case, we need to determine what to use in place of the missing data.  Let the measurement vector be partitioned as $z=(\tilde{z},z')$ where $z'$ is the missing data.  The Bayesian estimator with missing $z'$ is given by $\hat{x}'(z) = \mbbE(x|\tilde{z})$.   Given the originally trained neural network $\Kc(z;w)$ that approximates $x^*(z)=\mbbE(x|z)$, we have
\[
\tcb{\hat{x}'(z) = \mbbE_{z'}(\mbbE(x|\tilde{z},z'))\approx
\int \Kc((\tilde{z},z');w) dF_{z'},}
\]
where $F_{z'}$ is the cumulative distribution of $z'$.  Thus the Bayesian state estimator with missing $z'$ is the state estimates (without missing data) averaged over the missing data, which  can be implemented by resampling missing measurements and computing the  averaged state estimates. The re-sampling, however, is costly.   A simple heuristic   is replacing $z'$ with estimated $\mbbE(z')$ obtained in the training process (\ref{eq:se2xz}).

Next, we consider the problem of detecting and identifying bad data that may be results of sensor malfunctions or cyber-attacks.   The detected bad data can then be removed, and the above solution to missing data problem  can be applied.

We treat bad data as outliers.  To this end, the Bayesian formulation offers a direct way to detect and identify bad data. In contrast to the conventional bad-data detection methods that are based on residue errors computed after state estimation, we propose a {\em pre-estimation} bad detection technique  by exploiting the learned (prior)  distributions of the measurements.

We formulate a binary hypothesis testing problem where hypothesis $\Hc_0$ models measurements without bad data and $\Hc_1$ for measurements with bad data.    Consider first the simpler case when, under $\Hc_0$,  measurement $Y$ is Gaussian with mean $\mu_0$ and variance $\sigma_0^2$.  Under $\Hc_1$, $Y$ has a different distribution with mean or variance unequal to those under $\Hc_0$.  This is a composite hypothesis testing problem for which the uniformly most powerful test may not exist.

A widely used practical scheme is the Wald test \cite{Wasserman:book1} that evaluate the normalized deviation of the measurement away from $\mu_0$; $\Hc_1$ (bad data) is declared when the deviation exceeds a certain threshold. Specifically, given $Y=y$, the size $\alpha$ Wald test is given by
\[
\Bigg|\frac{y-\mu_0}{\sigma_0}\Bigg| \begin{array}{c} \Hc_1\\ \gtrless  \\ \Hc_0\\\end{array} z_{\alpha/2}:=Q^{-1}(\alpha/2),
\]
where $Q(x)=\frac{1}{\sqrt{2\pi}} \int_x^\infty \exp(-u^2/2)du$ is the Gaussian tail probability function. Typically, the size parameter $\alpha$ is set at $\alpha=0.05$ to ensure that the false alarm (false positive) probability is no greater than $5\%.$

For the application at hand, the mean $\mu_0$ and variance $\sigma_0^2$ of the measurement under $\Hc_0$ (no bad data) used in the Wald test are learned as a by-product of the training process described in Sec~\ref{sec:III}. Specifically, the solutions of the power flow equations give directly samples of the measurements, from which the mean and variance can be estimated.

Strictly speaking, however, the measurement distributions are not Gaussian, and an alternative to the Wald test can be derived by using explicitly the learned distribution.  This appears to be unnecessary from our simulation examples shown in Section~\ref{sec:V}.

%% file: simulationArxFinal.tex
\subsection{Simulation Settings}
\paragraph{Systems simulated}  The simulations were performed on two systems defined in the MATPOWER toolbox \cite{Zimmerman:11RZ}. One is an 85-bus low voltage system, the other one is the 3120-bus ``Polish network''. The simulation of the larger network was to demonstrate the scalability of the  proposed technique and its application to mesh networks.

Two types of measurement devices were assumed: (i) Current magnitude meters were placed in 20$\%$ of distribution system branches. (ii) One SCADA meter was placed  at the slack bus to measure complex power to/from the transmission grid. The additive measurement noise was assumed to be independent and identically distributed Gaussian  with zero mean and variance set at $1\%$ of the average net consumption value.

\paragraph{Performance measure}  The performance of the tested algorithms was measured by the per-node average squared error (ASE)  defined by
\begin{equation} \label{eq:ASE}
\mbox{ASE}=\frac{1}{MN}\sum_k ||\hat{x}[k]-x[k]||^2,
\end{equation}
where $M$ is number of Monte Carlo runs, $k$ the index of the Monte Carlo run, $N$ the number of nodes, $\hat{x}[k]$ and $x[k]$ the estimated and the state vectors, respectively.

\paragraph{Neural network specification and training}
The inputs of the neural network were current magnitudes and slack bus measurements; the outputs were the state estimates. The ReLU (Rectified Linear Units) activation function was used for neurons in the hidden layers and  linear activation functions in the output layer.

The Adam algorithm  \cite{Kingma&Ba:15ICLR}  was used to train the neural network with mini batches of 32 samples. Early stopping was applied by monitoring validation errors. To select an initial point for the optimization, He's normal   method  \cite{He:15CORR} was used.

\subsection{Distribution learning}
We used data sets from the Pecan Street collection\footnote{\url{http://www.pecanstreet.org/}} for distribution learning.
Specifically, a data set covering May 21st to September 21st of 2015 was used for training, and a data set covering the same period in 2016 was used for testing.  The data sets included power measurements from 82 households, from which 55 had solar PV installed. Power consumptions and solar PV generations were measured separately.  Distributions of active power consumptions and solar generations were learned from data separately.

\begin{figure}[h]
	\centering
	\includegraphics[width=1\columnwidth]{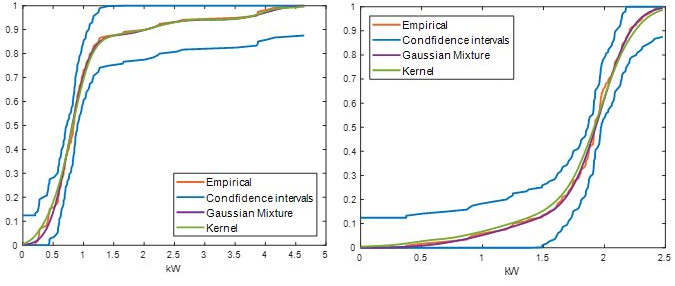}
	\vspace{-1em}
	\caption{\small Estimated cumulative distribution and 95\% confidence interval. }
	\label{fig:Distributions}
\end{figure}

We considered non-parametric (histogram and Kernel) and parametric distribution estimation techniques.  The latter included Gaussian, Gaussian mixture, and Weibull models.  Fig.~\ref{fig:Distributions} shows a representative of the cumulative distribution estimates of consumption and solar generation at 3 p.m. on bus 12 based on the histogram, 3-component Gaussian mixture, Kernel Epanechnikov estimators, and the 95$\%$ confidence bounds.  These estimates produced similar results for all distributions and were well within the non-parametric confidence bounds obtained using the Dvoretzky-Kiefer-Wolfowitz inequality \cite{Wasserman:book1}. For all cases, 100$\%$ of the 3-component Gaussian mixture estimates were within the 96$\%$ level confidence bound, and 99.8$\%$ of all estimates were within the 85$\%$ level confidence bound. We observed that Gaussian and Weibull distributions exceeded the 95$\%$ confidence interval that captures the true distribution. Our data analysis led to the adoption of Gaussian mixture model with three components in simulations.

\vspace{-0.5em}
\subsection{Simulation Results for a 85-bus radial  network}
\begin{figure}[!h]
	\centering
	\includegraphics[width=0.7\columnwidth]{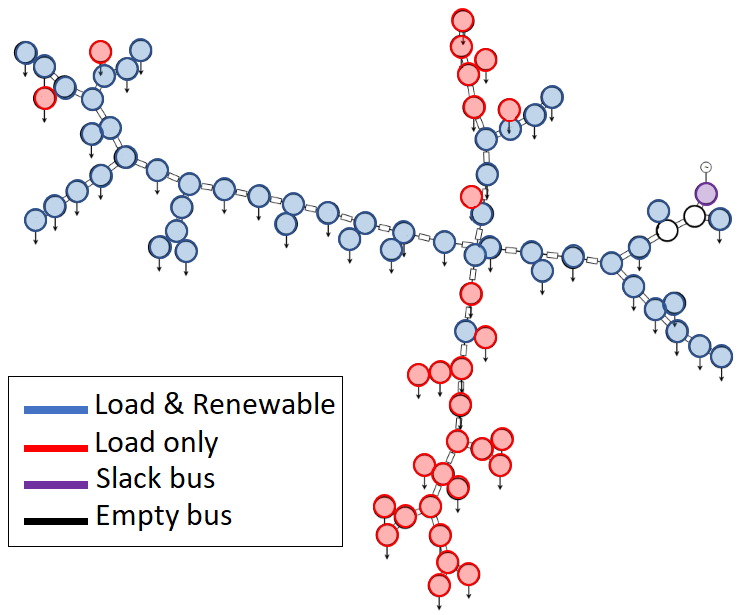}
\vspace{-1em}
	\caption{\small 85 bus system.}
	\label{fig:85busses}
\end{figure}

For the 85-bus network, 82 buses were assumed to have consumptions, and  55 arbitrarily chosen buses had renewable generations. The network configuration is presented in Fig~\ref{fig:85busses}. Renewable injections and load were generated separately to produce net injections.  We evaluated the ASE performance as defined in (\ref{eq:ASE}), CPU time,  and the performance of bad-data detection/filtering.

\paragraph{ASE  performance}
We compared the proposed  {\em Bayesian state estimation with deep neural network} (herein abbreviated as BSEdnn) with two  WLS-based pseudo-measurement methods in the literature:
\ben

\item  WLSp:  referred to as  {\em WLS with pseudo measurements}, WLSp generates injection pseudo measurements by  averaging the energy consumption measurement over several past samples \cite{Angioni&Schlosser&Ponci&Monti:16TIM};

\item WLSnnp: referred to as {\em WLS with neural-network generated pseudo measurements}, WLSnnp uses  a neural network to generates pseudo-measurements of net injection based on a regression on the last energy consumption vectors   \cite{Manitsas&Singh&Pal&Strbac:12TPS}.

\een

Separate neural networks were implemented and trained for each hour of the day.  We generated 10,000 training and 10,000 validation samples. The training of the networks took on average 500 update iterations.

\begin{figure}[!h]
	\centering
	\includegraphics[width=1\columnwidth]{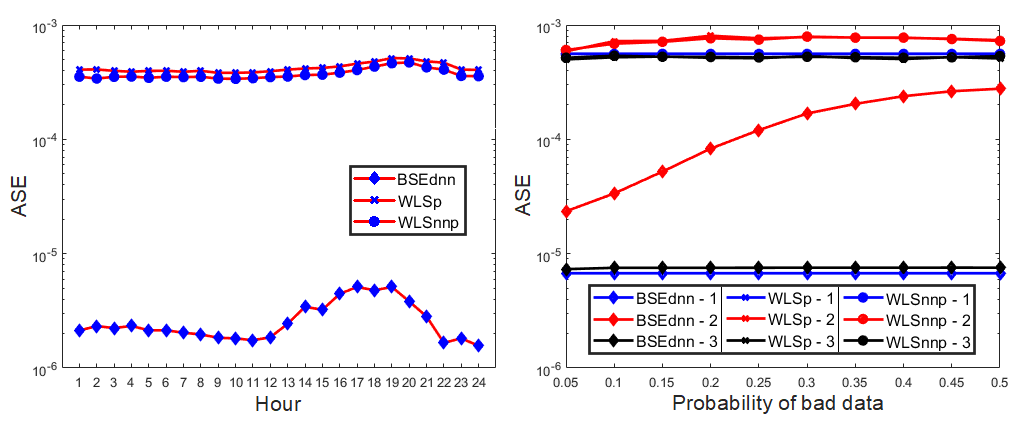}
\vspace{-1em}
	\caption{\small ASE performance for three state estimators. Left: ASE (in absence of bad data) of estimators in different hours. Right: ASE at hour 17 in the presence of bad data of different strength; Line 1 shows ASE without bad data, Line 2 shows ASE with bad data and Line 3 shows the ASE with bad-data detection and filtering for estimators.}
	\label{fig:ASE}
\end{figure}

Fig~\ref{fig:ASE} (left) shows that the ASE performance of the three state estimators on the test data for the 24 hours period. Across all hours, BSEdnn performed significantly better than the two pseudo-measurement-based WLS techniques. Specifically, the ASE of BSEdnn achieved one to two orders of magnitude lower ASE than the pseudo-measurement techniques.  The performance gain was attributed to that, although measurement distributions were in some way encoded in the pseudo-measurements, WLS with pseudo measurements misused this information.

\paragraph{CPU time} The experiments were carried out on a computer Intel Core i7-8700 with 3.2 GHz processor and 64 GB RAM. For the simulation result presented in Fig.~\ref{fig:ASE}, the computation time of the three algorithms was measured. Estimating states using WLS with pseudo measurements took on average 22.515 seconds per Monte Carlo run.  In contrast, calculating state estimate after the neural network had been trained took on average only 0.16 millisecond per Monte Carlo run, which was five orders of magnitude faster than the WLS estimator.  The training algorithm was implemented in Python 3.6.5 using Keras v2.2.4 with Tensorflow v 1.12 as the backend [54]. Training the network with five layers and 2000 neurons took 23 minutes and 7 seconds.

\paragraph{Bad-data detection and filtering}  The performance of bad data-detection and filtering was tested.  In a Monte Carlo simulation with 1,000 runs,    bad data were injected to measurements probabilistically; each measurement had probability $\eta$ to contain bad data.  Different  strengths of the bad data were evaluated by varying $\eta$.  Note that the way bad data were introduced deviated from and  significantly more challenging than  the conventional settings where a fixed number of bad data at specific locations was typically assumed.

For the bad-data model considered in Sec~\ref{sec:IV}, we assumed that, under hypothesis $\Hc_1$ (bad data), the additive noise was Gaussian with zero mean and considerably larger standard deviation $\sigma_1$ than $\sigma_0$ under $\Hc_0$ (no bad data)\footnote{Note that neither the mean and the standard deviation nor the distribution of the bad data were assumed known in the bad-data detection and filtering algorithm.}.
Once bad data were detected, BSEdnn performed a data filtering procedure that replaced the bad data with the mean of the measurements as described in Sec~\ref{sec:IV}. For WLS-p and WLS-nnp, detected bad data were removed from the WLS procedure.

 Fig~\ref{fig:ASE} (right) shows the ASE performance vs. the probability $\eta$ of bad-data occurrence under the bad-data model  $\sigma_1=10\sigma_0$. The bad-data detection and filtering algorithms were compared in three cases: Case 1 shown in blue lines  were the baseline performance without bad data; Case 2 shown in red lines were the performance when bad data were present but not filtered.  Case 3 shown in black was for the performance with bad-data filtering.  It was evident that bad-data detection and filtering improved the performance  of state estimation considerably, driving the ASE performances  closer to those for the clean data case. It should also be observed that, even without bad-data filtering,  BSEdnn performed better than pseudo-measurement based WLS estimators.

Additional simulations were conducted to evaluate the performance of bad-data detection. Recall that, in deriving the threshold used in the Wald test for bad-data detection in Sec~\ref{sec:IV},  a Gaussian distribution of the measurement under $\Hc_0$ was assumed, which at best could only be an approximation given the Gaussian mixture nature of the net injection.  To evaluate the performance of bad detection derived from the Gaussian model, we provided a comparison with the case when noise distributions were indeed Gaussian, for which the theoretical value of false alarm and miss detection probabilities could be computed analytically.

Table~\ref{table:Bad} shows bad-data detection and filtering performance. The top two blocks of the table show that the threshold obtained based on the Gaussian model approximated well the false alarm and detection probabilities under the Gaussian mixture models. Also shown here is that the pseudo-measurement schemes with WLS did not perform well. The third, fourth and fifth blocks show the effects of bad data on the ASE of state estimators. The bad-data filtering algorithm mitigated the effects of bad data satisfactorily.

\begin{table}[ht]
	\centering
	\scalebox{0.7}{\begin{tabular}{||l |c c c||} 
			\hline\hline 
			& $\sigma_1=5\sigma_{0}$ & $\sigma_1=10\sigma_{0}$ & $\sigma_1=20\sigma_{0}$  \\ [0.5ex]
			\hline\hline
			False alarm prob. (Gaussian theoretical) 	& 5.00\%	 	& 5.00\%	 		& 5.00\%	 	\\
			False alarm rate of Wald test				& 4.60\%		& 4.70\%			& 4.70\%		\\
			False alarm rate of WLSp+J(x)				& 4.94\%		& 4.78\%			& 4.91\%		\\
			False alarm rate of  WLSnnp+J(x)			& 4.94\%		& 4.78\%			& 4.91\%		\\\hline
			Detection prob. (Gaussian Theoretical)		& 69.5\%		& 84.5\%			& 92.1\%		\\
			Detection rate of Wald test					& 68.14\%		& 84.16\%			& 91.73\%		\\
			Detection rate of WLSp+J(x)					& 36.45\%		& 41.75\%			& 57.33\%		\\
			Detection rate of WLSnnp+J(x)				& 36.75\%		& 43.37\%			& 58.96\%		\\\hline\hline
			ASE of BSEdnn without bad data				& 6.50E-06	  	& 6.50E-06	    	& 6.50E-06  	\\
			ASE of WLSp  without bad data				& 5.87E-04	  	& 5.87E-04	    	& 5.87E-04  	\\
			ASE of WLSnnp MSE without bad data			& 5.85E-04	  	& 5.85E-04	    	& 5.85E-04   	\\\hline
			ASE of BSEdnn with bad data					& 1.84E-05	  	& 9.01E-05		    & 7.31E-04     	\\
			ASE of WLSp with bad data	 				& 6.97E-04	  	& 7.06E-04	  	    & 7.75E-04     	\\
			ASE of WLSnnp with bad data 				& 6.98E-04	  	& 6.90E-04  	    & 7.70E-03  	\\\hline
			ASE of BSEdnn with bad-data filtering		& 6.91E-06	  	& 6.91E-06		    & 6.86E-06 		\\
			ASE of WLSp  with bad-data filtering 		& 5.82E-04	  	& 5.89E-04	    	& 5.79E-04		\\
			ASE of WLSnnp MSE with bad-data filtering	& 5.76E-04	  	& 5.82E-04	    	& 5.79E-04   	\\[1ex]
			\hline\hline 
	\end{tabular}}
	\caption{Bad-data detection and filtering with varying strength of bad data. $\eta = 0.3$}
	\label{table:Bad}
\end{table}

\vspace{-0.5em}

\vspace{-0.5em}
\subsection{Optimizing the Structure of Deep Neural Network}
 We examined the effects of choosing different structures of the neural network on the performance of BSEdnn. To this end, we considered the two questions: (i) Given a fixed number of neurons, is there an optimal depth of the neural network that offers the best performance? (ii) Is there a particular ``shape'' of the network that should be favored?  We attempted to address these issues through simulations, knowing that interpretations presented here  apply to the system studied and may not be conclusive.

\begin{figure}[h]
	\centering
	\includegraphics[width=0.7\columnwidth]{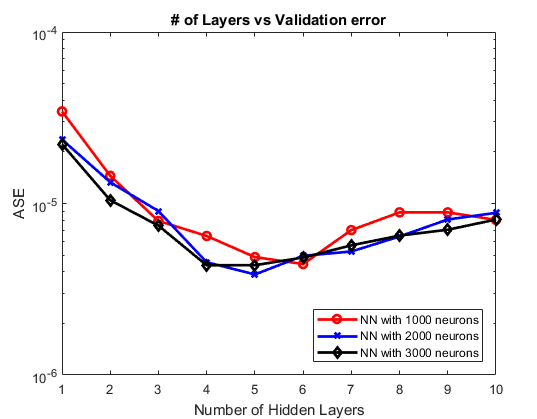}
	\hphantom{lol}
	\vspace{-0.5em}
	\caption{\small NN Architecture - Total number of neurons fixed.}
	\label{fig:Architecture1}
\end{figure}

\paragraph{The depth of deep learning}
For total 1000, 2000, and 3000 neurons, we  tested exhaustively different rectangular neural networks of different depths until a trend was observed. The ASE performance of BSEdnn using validation data sets is shown in   Fig~\ref{fig:Architecture1}. The results suggested that a neural network of 4 to 6 layers perform the best.

\paragraph{The shape of the deep neural network}  Next we examined whether using an irregular shaped neural network has an advantage. To this end, we considered a pruning approach that, starting from a rectangular network, progressively combining neurons that were highly correlated based on the clustering technique developed in Sec~\ref{sec:III}.  The idea was that, by combining neurons whose outputs were highly correlated, the resulting irregular neural network would be regularized within a particular structure that, potentially, could lead to faster convergence and better performance.

Table~\ref{table:Cluster} summarizes the simulation results. First, we evaluated the rectangular structure with 400 neurons in each layer. Second, we used the clustering algorithm to prune the network, which resulted in a reduction of the number of neurons at the two ends of the network and improved test error.  The second round of clustering-pruning resulted in further reduction training, test, and validation errors, and a shape of the network with the second and third  layers having considerably more neurons than the first and last two layers.  The third round of clustering-pruning resulted in increased errors, indicating the locally best clustering-pruning procedure should end at the second iteration.

\begin{table}[ht]
	\centering
	\scalebox{0.7}{\begin{tabular}{||l | c c c c||}
			\hline\hline
			
			& Rectangular		 & 1st Pruning 	  & 2nd Pruning		  & 3rd Pruning \\ [0.5ex]
			\hline
			\# of Neurons 1st layer		& 400				& 399			& 395			& 391			\\
			\#  of Neurons 2nd layer	& 400				& 400			& 400			& 400			\\
			\#  of Neurons 3rd layer	& 400				& 400			& 399			& 380			\\
			\#  of Neurons 4th layer	& 400				& 346			& 278			& 173			\\
			\#  of Neurons 5th layer	& 400				& 154			& 64			& 22			
			\\\hline
			Test error		    		& 6.23e-06 	& 5.77e-06 		& 4.45e-06	& 5.11e-06	\\
			Validation error			& 4.73e-06	& 4.64e-06		& 3.72e-06	& 3.85e-06	\\
			Training error				& 4.27e-06	& 4.13e-06		& 3.41e-06	& 3.73e-06	\\ [1ex]
			\hline
	\end{tabular}}
	\caption{Pruning simulation results with the threshold = 0.0005}
	\label{table:Cluster}
\end{table}

\subsection{Simulation results for a 3120-bus mesh network}
To demonstrate that the proposed algorithm can scale computationally for a larger network, and that the developed scheme also applies to a general mesh network,  we applied our approach on the 3120-bus mesh network. For this "Polish" network, 2664 of the busses were assumed to have stochastic consumption, and 348 of the busses having stochastic generations.

In general, transmission systems are observable.  To illustrate the idea of state estimation in an unobservable system, we
assumed that only 20\%  of the power injection measurements are available. A  3-layer network with 1000 neurons in each layer is trained.  Again, the neural network was trained with data generated from one injection distribution and tested with a different injection distribution.  The injection distributions were the same used in the simulations for the 85-bus network.

We considered three test cases:
\ben
\item Normal data: power injection measurements had nominal measurement noise with standard deviation of $1\%$ of the average net consumption value.
\item  Bad data:  30\% of the measurements contained  bad data modeled as having measurement noise ten times of the  nominal standard deviation.
\item Missing data: 30\%  of the data were missing.
\een

\begin{table}[ht]
	\centering
	\scalebox{0.7}{\begin{tabular}{||l |c c c c||} 
			\hline\hline 
			& Training & Normal & Bad data & Missing data   \\ [0.5ex]
			\hline\hline
			ASE 	        & 2.569e-07  & 8.962e-06	 	& 1.082e-05 		& 1.470e-05	 	\\
			Comp. Time	& 2 hours and 33 minutes  & 0.377 ms	  	& 0.358 ms	    	& 0.350 ms  	\\[1ex]
			\hline\hline 
	\end{tabular}}
	\caption{\small 3120 bus system simulation results.  {40,000 samples were used in training and 20,000
Monte Carlo runs were used  in testing.}}
	\label{table:3120}
\end{table}

Table~\ref{table:3120} shows the training and testing errors for the three scenarios.  While standard nonlinear programming routines in MATLAB failed for the network of this size, the Bayesian state estimated implemented by a deep neural network performed well.  The training and testing errors for the three cases achieved a similar performance as the 85-bus system.
This experiment showed that deep neural network required only millisecond level time to estimate states of the 3120 bus system.   Training took more than 2 hours, however, which means that the network needed to be trained offline several hours ahead of the time of operation. 

%% file: appendixArxFinal.tex
\subsection{Three-Phase Unbalanced Power Flow }\label{sec:3P}
We derive here the functional relationship between measurements and states. Different configurations of measurements can be assumed. Possible measurements that make of the measurement vector $z$ include, for each phase $k \in \{1,2,3\}$,
\begin{tabular}{@{}p{.15cm}ll}
${P_{i}^{k}}$   & : &   Active power injection at node $i$;\\
${Q_{i}^{k}}$   & : &  Reactive power injection at node $i$;\\
${P_{ij}^{k}}$  & : &   Active power flow from node $i$ to $j$;\\
${Q_{ij}^{k}}$  & : &   Reactive power flow from node $i$ to $j$;\\
${I_{ij}^{k}}$  & : &   Current from node $i$ to $j$.\\
\end{tabular}


For distribution systems, a (small) subset of the power flow variable set $\Zmsc= \{P_i^k, Q_i^k, P_{ij}^k, Q_{ij}^k, |I_{ij}^k|\}$ are made available, where $i,j$ are bus indices and $k$ the phase index.   The real and reactive power injections are related to the state variables by

\begin{equation}\label{active_injection}
P_{i}^{k}=V_{i}^{k} \sum_{l=1}^{3} \sum_{j=1}^{n} V_{i}^{l}[G_{ij}^{kl}\cos(\theta_{i}^{k}-\theta_{j}^{l})+B_{ij}^{kl}\sin(\theta_{i}^{k}-\theta_{j}^{l})],
\end{equation}
\hphantom{lol}
\vspace{-0.3cm}
\begin{equation}\label{reactive_injection}
Q_{i}^{k}=V_{i}^{k} \sum_{l=1}^{3} \sum_{j=1}^{n} V_{i}^{l}[G_{ij}^{kl}\sin(\theta_{i}^{k}-\theta_{j}^{l})+B_{ij}^{kl}\cos(\theta_{i}^{k}-\theta_{j}^{l})],
\end{equation}

\noindent where $G_{ij}^{kl}$ and $B_{ij}^{kl}$ are the conductance and susceptance between node $i$ and $j$ from phase $k$ to $l$. The branch power flows $P_{ij}^k$ and $Q_{ij}^k$ are related to the state variables by

\vspace{-0.1cm}
\setlength{\arraycolsep}{0.0em}
\begin{eqnarray}
\vspace{\medskipamount=20pt}
\setlength{\abovedisplayshortskip}{-12pt}
\setlength{\belowdisplayshortskip}{-1pt}
P_{ij}^{k}&{}={}&V_{i}^{k} \sum_{l=1}^{3} V_{i}^{l}[G_{ij}^{kl}\cos(\theta_{i}^{k}-\theta_{j}^{l})+B_{ij}^{kl}\sin(\theta_{i}^{k}-\theta_{j}^{l})]\nonumber\\\vspace{-0.3cm}
&&\:{-}V_{i}^{k} \sum_{l=1}^{3} V_{j}^{l}[G_{ij}^{kl}\cos(\theta_{i}^{k}-\theta_{j}^{l})+B_{ij}^{pq}\sin(\theta_{i}^{k}-\theta_{j}^{l})],\nonumber\\*
\end{eqnarray}
\hphantom{lol}
\vspace{-1.0cm}
\setlength{\arraycolsep}{0.0em}
\begin{eqnarray}
Q_{ij}^{k}&{}={}&-V_{i}^{k} \sum_{l=1}^{3} V_{i}^{l}[G_{ij}^{kl}\sin(\theta_{i}^{k}-\theta_{j}^{l})-B_{ij}^{kl}\cos(\theta_{i}^{k}-\theta_{j}^{l})]\nonumber\\ \vspace{-0.3cm}
&&{-}\:V_{i}^{k} \sum_{l=1}^{3} V_{j}^{l}[G_{ij}^{kl}\sin(\theta_{i}^{k}-\theta_{j}^{l})-B_{ij}^{kl}\cos(\theta_{i}^{k}-\theta_{j}^{l})].\nonumber\\*
\end{eqnarray}

\hphantom{lol}
\vspace{-0.6cm}

Current magnitude measurements $|I_{ij}^k|$ are often included in the system as additional measurements \cite{Abur&Exposito:97TPS}. They can be used to strengthen state estimation and they are related to state variables by

\vspace{-0.3cm}
\begin{eqnarray}\label{currentReal}{
\vspace{-2\baselineskip}
}
\operatorname{Re}(I_{ij}^{k})&{}={}&\sum_{l=1}^{3} V_{i}^{l}[G_{ij}^{kl}\sin(\theta_{i}^{l})-B_{ij}^{kl}\cos(\theta_{i}^{l})]\nonumber\\\vspace{-0.4cm}
&&{-}\:\sum_{l=1}^{3} V_{j}^{l}[G_{ij}^{kl}\sin(\theta_{j}^{l})-B_{ij}^{kl}\cos(\theta_{j}^{l})],
\phantom{\hspace{1.5cm}}
\end{eqnarray}
\vspace{-0.4cm}
\begin{eqnarray}\label{currentImaginary}{
\vspace{-2\baselineskip}
}
\operatorname{Im}(I_{ij}^{k})&{}={}&{-}\sum_{l=1}^{3} V_{i}^{l}[G_{ij}^{kl}\sin(\theta_{i}^{l})+B_{ij}^{kl}cos(\theta_{i}^{l})]\nonumber\\\vspace{-0.4cm}
&&{-}\:\sum_{l=1}^{3} V_{j}^{l}[G_{ij}^{kl}\sin(\theta_{j}^{l})+B_{ij}^{kl}\cos(\theta_{j}^{l})].
\phantom{\hspace{1.5cm}}
\end{eqnarray}

The energy consumption measurements are modeled as the accumulated values of the power injections $P_{i}^k$.

\subsection{Input-Output Relation of Deep Neural Networks} \label{sec:NN}

For an $L$ layer neural network, the first $L-1$ layers are hidden layers and the last layer the output layer.   The input $u_{i}^{(k)}$ of  neuron $i$ in  hidden layer $k$ is an affine combination of the outputs   of the $(k-1)$th layer
\beq \label{eq:W}
u_{i}^{k}=w_{i,0}^{(k)} + \sum_{j=1} w_{i,j}^{(k)} z_{j}^{(k-1)},~~z_{i}^{(k)}=\rho(u_i^{(k)}),
\eeq
where $z_i^{(k)}$ is the output of the $i$th neuron in layer $k$,  $\rho(\cdot)$ the so-called activation function, and $\{w_{i,j}^{(k-1)}\}$ is the set of the weighting coefficients for the outputs from neurons at  layer $(k-1)$.    The activation function $\rho$ can be chosen in many ways; some of the common choices include the biologically inspired  sigmoid function family and various forms of  rectifier functions.

In a vector-matrix form, the output of vector $z^{(k)}$ of hidden layer $k$ is related to the output vector  $z^{(k-1)}$ of hidden layer $(k-1)$  by
\[
z^{(k)} = \rho(\Wc^{(k)}(z^{(k-1)})),
\]
where $\Wc^{(k)}(\cdot)$ is an affine mapping defined in (\ref{eq:W}) from the output of layer $(k-1)$ to the input of layer $k$.  Here  we adopt the convention that, for a vector $x$, $\rho(x)=[\rho(x_1),\cdots,\rho(x_N)]^{\intercal}$.

The overall intput-output relation of the multi-layer neural network is therefore an $L$-fold iterated nonlinear map from the measurements as the input to state estimates as the output
\bea
\hat{x} &=& w^{(L)}_0+\sum_{j=1}^N w_j^{(L)} \rho(\Wc^{(L-1)}(\rho(\cdots \rho(\Wc^{(1)}(z)))))\nn\\
&:=& \Kc(w,z), \label{eq:K}
\eea
where $w$ is a matrix that includes all $w^{(k)}_{i,j}$ as its entries.

\subsection{Injection Distribution Learning from Smart Meter Data} \label{sec:smartmeter}

We present here a technique that estimates the power injection distribution using data collected from smart meters.  The main technical challenge is that  smart meters measure the {\em cumulative energy consumption (kWh)} at a timescale  slower than that of the SCADA measurements.

For convenience, assume that the fast timescale data (SCADA) are produced every interval of unit duration  whereas the smart meter measurements
are produced once every $T$ intervals. Let the sequence of the smart meter measurements be $Y_t$.  The $t$th interval has $T$ fast-time measurements $X_{t1},\cdots, X_{tT}$.  We have
\beq\label{eq:YtXt}
Y_t= \sum_{n=1}^T X_{tn}.
\eeq
Here we make the assumption that $(X_{ti})$ is a stationary process which makes $Y_t$ also stationary.

Empirical studies have suggested that a good model for the probability density function (PDF) $f_Y$ of $Y_t$ is a Gaussian mixture, \ie
$$f_Y = \sum_{i=1}^K \pi_i \Nc(m_i,V_i^2),$$ where $\Nc(m,V^2)$ denotes the Gaussian distribution with mean $m$ and covariance $V^2$.   Such a model has the interpretation that there are $K$ operating patterns (such as weather or the tune if use), each can be modeled by a Gaussian process. With probability $\alpha_i$, the $i$th pattern shows up in the interval of measurement.   Under this interpretation, it is reasonable to assume that $X_{ti}$ also follows the same underlying physical model: the distribution $f_X$ of $X_{ti}$ is also a Gaussian mixture with the same number of components and the same probability of appearance, \ie $f_X = \sum_{i=1}^K \pi_i \Nc(\mu_i, \sigma_i^2).$
Therefore,  we only need to find $\mu_i$ from $m_i$ and $\sigma_i^2$ from $V_i^2$.

We now fix $t$ and a particular Gaussian mode $i$ to derive a procedure to compute $\sigma_i^2$ from $V_i^2$.   To avoid cumbersome notations,
we drop the component index $i$ in the Gaussian mixture and the subscript $t$, writing  $X_{tn}$ as $X_n$, $V_i^2$ as $V^2$, and $\sigma_i^2$ as $\sigma^2$.

From (\ref{eq:YtXt}),  we have immediately $\mu_i = \frac{1}{T} m_i$. Let  $C_x(k)$ be the  auto-covariance of $X_n$ with lag $k$.  Then, the variance $V^2$ of $Y_t$ satisfies
\bea
V^2 &=& \sum_{i,j=-T+1}^{T-1}C_x(i-j) = \gamma^{\T}c  \label{eq:V}\\
c&:=&[C_x(0),\cdots, C_x(T-1)]^{\T}, \nn\\
\gamma &:=& [T, 2(T-1), 2(T-2),\cdots, 2]^{\T}. \nn
\eea
Our goal is to obtain $\sigma^2=C_x(0)$ from $V^2$.

We now make the assumption that $X_n$ is a stationary AR-$K$ process defined by
\bea \label{eq:ARM}
X_n=\alpha_1X_{n-1}+\cdots+\alpha_K X_{n-K} +  \epsilon_n,
\eea
where $\epsilon_n \in \Nc(\mu_\epsilon,\sigma_\epsilon^2)$ is the innovation sequence that is IID Gaussian. With a one or more traces of $X_n$, the AR parameters $\alpha=(\alpha_k)$ can be easily estimated using, \eg the least squares method.

Knowing $\alpha$, the auto-covariance functions can be computed directly from a variation of the Yule-Walker equation:
\[
c=A_\alpha c+\sigma_\epsilon^2 e_1~~\Rightarrow~~c=\sigma^2_\epsilon (I-A_\alpha)^{-1} e_1,
\]
where $e_1=[1,0,\cdots, 0]^{\T}$, and $A_\alpha$ is a matrix made of entries of $\alpha$.  Substituting $c$ into (\ref{eq:V}), we obtain
\[
\sigma^2= \frac{e_1^{\T} (I-A_\alpha)^{-1} e_1 }{\gamma^{\T} (I-A_\alpha)^{-1} e_1} V^2.
\]
To summarize, from the smart meter data $Y_t$, we obtain first  the Gaussian mixture coefficients $\{(\pi_i,  m_i, V_i^2)\}$, which gives the coefficients $\{(\pi,\mu_i,\sigma_i^2)\}$  of the Gaussian mixture of $X_{tn}$.  Specifically,
\bea
\mu_i &=&  \frac{1}{T} m_i,~~\sigma^2_i = \frac{e_1^{\T} (I-A_\alpha^{(i)})^{-1} e_1}{ \gamma^{\T} (I-A_{\alpha^{(i)}})^{-1} e_1} V_i^2,
\eea
where $\alpha^{(i)}$ is the coefficients of the AR process associated with component $i$.